# Deep Convolution Network Based Emotion Analysis for Automatic Detection of Mild Cognitive Impairment in the Elderly

Zixiang Fei[1], Erfu Yang[1]*, Leijian Yu[1], Xia Li[2], Huiyu Zhou[3], Wenju Zhou[4$]

[1] Department of Design, Manufacturing and Engineering Management
University of Strathclyde, Glasgow G1 1XJ, UK
zxfei89n@163.com, {erfu.yang, leijian.yu}@strath.ac.uk
[2] Shanghai Jiaotong University, Shanghai, China
lixia11111@alumni.sjtu.edu.cn
[3] Department of Informatics
University of Leicester, Leicester LE1 7RH, UK
hz143@leicester.ac.uk
[4] Shanghai University, Shanghai, China
zhouwenju@shu.edu.cn

\* Correspondence: erfu.yang@strath.ac.uk; Tel.: +44-141-574-5279
$ Correspondence author in China: zhouwenju@shu.edu.cn; Tel.: +86-131-5692-9148

**Abstract:** A significant number of people are suffering from cognitive impairment all over the world. Early detection of cognitive impairment is of great importance to both patients and caregivers. However, existing approaches have their shortages, such as time consumption and financial expenses involved in clinics and the neuroimaging stage. It has been found that patients with cognitive impairment show abnormal emotion patterns. In this paper, we present a novel deep convolution network-based system to detect the cognitive impairment through the analysis of the evolution of facial emotions while participants are watching designed video stimuli. In our proposed system, a novel facial expression recognition algorithm is developed using layers from MobileNet and Support Vector Machine (SVM), which showed satisfactory performance in 3 datasets. To verify the proposed system in detecting cognitive impairment, 61 elderly people including patients with cognitive impairment and healthy people as a control group have been invited to participate in the experiments and a dataset was built accordingly. With this dataset, the proposed system has successfully achieved the detection accuracy of 73.3%.

## 1 Introduction

A large number of elderly people are suffering from dementia worldwide. People who have dementia may have worse memories and language disabilities, compared to normal people [1]. Significant money and workload are spent to take care of the people with dementia. As Mild Cognitive Impairment (MCI) can be considered an intermediate stage before dementia, early detection of MCI can be of great importance to both patients and caregivers.

However, current approaches for detecting MCI have their problems. For example, for traditional cognitive tests, neurophysiologists are required to carry out tests [2]. Furthermore, the neuroimaging techniques such as magnetic resonance imaging (MRI) may be used to diagnosis the disease, but will result in high expenses [3]. For example, some researchers proposed deep belief network-based frameworks for the detection and classification of MCI and Alzheimer's disease using magnetic resonance imaging [4]. In particular, the overfitting problem is solved by dropout technology and zero-masking strategy. The principal component analysis is utilized to reduce the feature dimension and the multi-task feature selection approach is also introduced. In a relevant research, the researchers proposed a framework for diagnosis of Alzheimer's disease using MRI image preprocessing, principal component analysis and the Support Vector Machine (SVM) [5]. To optimize the SVM parameters, a new switching delayed particle swarm optimization (SDPSO) algorithm is proposed.

On the other hand, patients with cognitive impairment are found to have abnormal emotion patterns which have the potential to work as an alternative method to detect the cognitive impairment [6][7][8]. For instance, Smith et al.'s research suggested that the patients with cognitive impairment showed more negative feelings when watching negative images [7]. Moreover, Henry et al. found that people with cognitive impairment had difficulties in facial muscle control [9].

In addition, there are no developed systems to detect cognitive impairment based on the analysis of emotions with good performance. Toward this end, this paper presents a new system that is designed for cognitive impairment detection in the early stage. There are three units in the system. The first unit is a user interface which is able to show the video stimuli to the participants and record their facial expressions at the same time. Next, the second unit is the proposed facial expression recognition system which can convert the video of facial expressions into a data matrix containing recognition results. Finally, the third unit is responsible for detecting the cognitive impairment of the participants based on the data matrix of facial expressions.

Computer vision techniques have been widely applied to many research areas including object detection, scripts analysis, and image classification [10–12]. Currently, automatic facial expression recognition is an important application of computer vision. Face detection, facial feature extraction and facial feature classification are three essential stages in automatic facial expression recognition. There are two categories of emotion analysis: static images-based and continuous video frames-based approaches. For instance, Jain et al. proposed a facial expression recognition algorithm based on a convolution neural network and static images, but the algorithm was only tested in two small datasets [13]. In Zeng et al.'s research, a novel framework for facial expression recognition was proposed using deep sparse autoencoders (DSAE) [14]. Geometric and appearance features are utilized as high-dimensional features in facial expression recognition. Also, DSAE is used to recognize facial expressions with a good performance by learning robust and discriminative features from the data. Dong et al. proposed a dynamic facial expression recognition algorithm [15]. In their deep learning-based algorithm, a shallow network was constructed to improve the accuracy when there was no sufficient training data.

The major contributions of our work are summarized below:

Firstly, a cognitive impairment detection system is proposed based on facial expression analysis with acceptable accuracy and low cost. The cognitive impairment is mainly detected by analyzing abnormal emotion patterns in a period of time.

Secondly, a novel facial expression recognition algorithm is proposed, which has good performance for people with cognitive impairment. The algorithm extracts the features using block_11_add layer from MobileNet [16] [17], and then uses Support Vector Machine (SVM) to classify these features [18].

Thirdly, a novel strategy in the experiment design is proposed, focusing on the facial expressions of the participants when they are watching the designed video stimuli. The experiments involving 61 participants are carried out to explore the abnormal emotion patterns related to cognitive impairment.

Furthermore, although some deep convolution networks such as AlexNet and ShuffleNet have good performance in the recognition of facial expressions, they are time-consuming. The proposed method (MobileNet + block_11_add + SVM) can solve this problem. Also, it only requires a single pass through the data [19]. Compared to MobileNet, the proposed method uses part of MobileNet for feature extraction and uses an SVM classifier to replace with the Softmax classifier in order that the operating time is reduced.

The organization of this paper is as follows: Section 1 gives an introduction to the research background, needs and motivations. The proposed facial expression system and cognitive impairment detection system is presented in Section 2. Section 3 mainly provides the experimental design and results. In Section 4, the detailed discussion and analysis on the experiment are carried out. Finally, the conclusion is given in Section 5.

## 2. Materials and Methods

*2.1 Overview of System Structure*

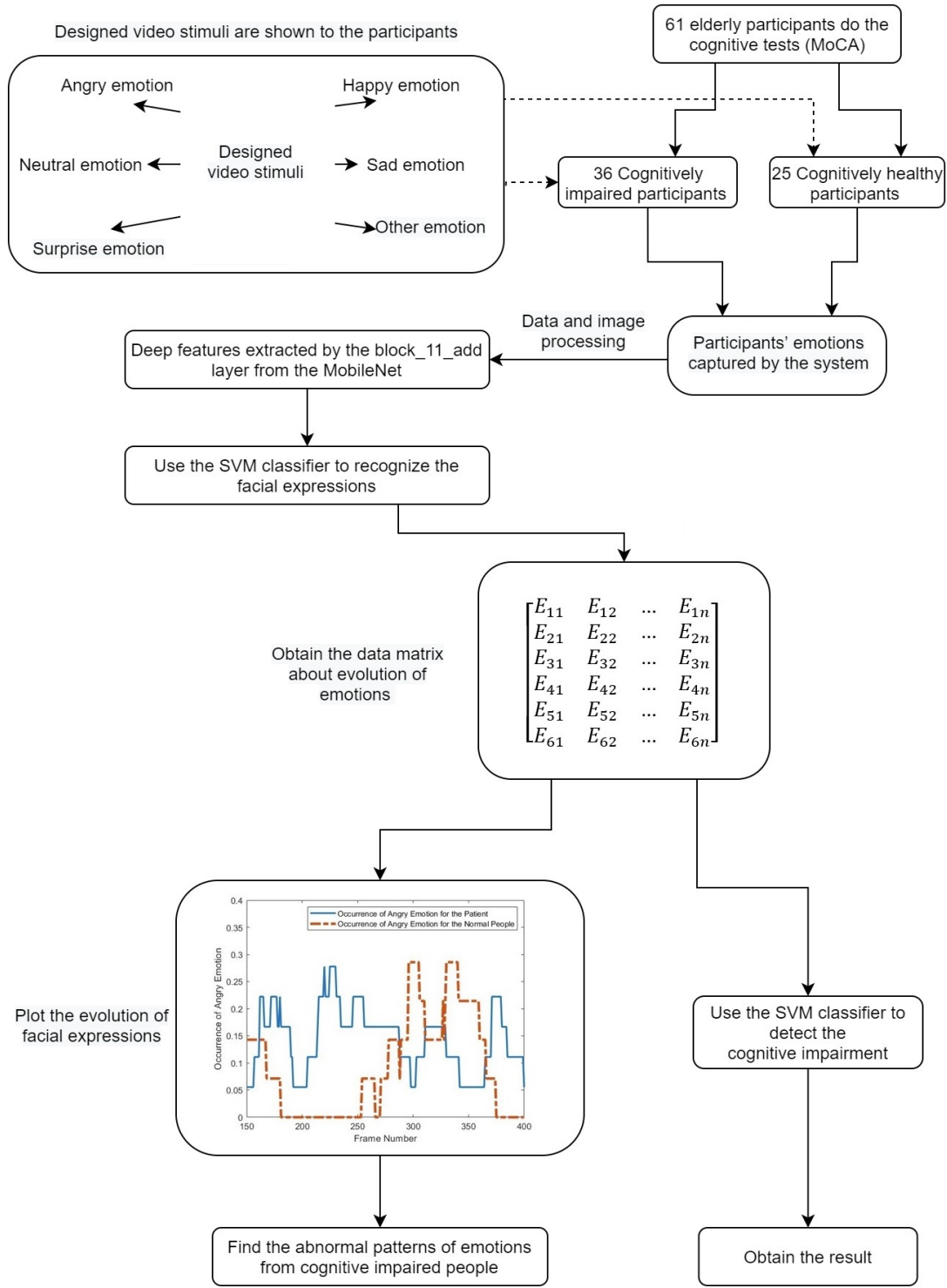

**Fig. 1.** Structure of the proposed system

For the early detection of cognitive impairment with high accuracy and low cost, a computer vision-based detection system is proposed. The major structure of the system is shown in Fig. 1. Three major parts are involved in this system: a developed interface to trigger the participants' emotions and record their facial expressions, a proposed facial expression recognition algorithm and a section to detect the cognitive impairment based on the recognized emotions.

In the experiment, a notebook computer was used to produce the video stimuli and record the facial expressions. 61 participants were invited to the experiments including 25 healthy people and 36 cognitively impaired participants. Our major process will be introduced in the following paragraphs. Before collecting the emotion data, the participants are asked to do the Montreal Cognitive Assessment (MoCA) which will be used to classify the participant group manually: cognitively impaired and cognitively healthy participants. Next, the developed interface will be utilized to display the designed video stimuli to the participants which will arouse participants' emotions including happy, neutral, sad, angry, surprise and other. Then, their reactions and facial expressions are recorded through the interface module.

After data processing, the proposed algorithm is used to recognize the facial expressions for each frame in the video for every participant: the layer block_11_add from MobileNet is used to extract the image features and the SVM trained by these deep features is used to recognize the emotions. As a result, for each participant, a data matrix showing his or her evolution of emotions during the experiment is created, which will be introduced in detail in Section 2.4.

Finally, two types of results are obtained. One type of result is about the emotion patterns. After the video is recognized frame by frame, the recognition result will be used to visualize the evolution of emotions of the cognitively impaired and healthy people, and the comparison will be introduced in Section 3.3. As a result, the cognitive impairment related abnormal emotion patterns can be found.

On the other hand, the SVM classifier is used to detect the cognitive impairment based on the data matrix about the evolution of emotions. The emotion data from 46 participants are used as the training dataset to train the SVM classifier and the emotion data from the remaining 15 participants are treated as the testing dataset.

*2.2 Proposed Facial Expression Recognition Algorithm*

The structure of the proposed facial expression recognition algorithm is shown in Fig. 2. The major sections in the structure are: the input, the image processing part, MobileNet to extract the image features, SVM to train and classify the extracted features and the output.

The first part of the proposed algorithm is the input. The videos recorded through the developed interface module are converted into continuous image frames as the input. The second part is about image processing to locate and crop the face part from the images and resize the images into the size 224×224 which is required by the MobileNet. The Viola-Jones algorithm is used to locate the position of the face which was proposed by Paul Viola and Michael Jones [20,21]. This method has some advantages such as good object detection rates and the ability to operate in real-time. As the proposed system needs to deal with large amounts of images of facial expressions, the advantage in operating time is important. After locating and cropping the face, the unnecessary image part like the background can be removed. By removing the unnecessary image part, the proposed emotion recognition algorithm will focus only on the human face, not the background environment which is important in the feature extraction stage.

Thirdly, the layer block_11_add from the pre-trained deep neural network MobileNet is used to extract the image features from input images. MobileNet is an efficient Convolution Neural Networks (CNN) which is tailored for mobile environments [16]. Convolution Neural Networks (CNN) often needs less image pre-processing before the feature extraction stage, and it is an artificial neural network that involves convolution operations [22]. There are many kinds of CNNs such as AlexNet which has 8 layers [23] and MobileNet which has 53 layers [17]. Furthermore, in previous research, it was found that combining part of CNNs with traditional classifiers like SVM and Linear Discriminant Analysis (LDA) can improve the recognition accuracy and reduce the operating time [24].

MobileNet has the advantage of reducing the number of operations and memory needed while retaining the same accuracy. MobileNet has been trained with the ImageNet dataset which has more than one million images in 1000 categories [25]. It has good performance in object classification, object detection and image segmentation.

MobileNet has different versions such as MobileNetV1and MobileNetV2 [16]. In this research, MobileNetV2 is adopted. They both use Depthwise Separable Convolutions to replace standard convolutions in order that the computational cost is reduced. As a result, the facial expression recognition system is able to operate on a device with low system requirements or a mobile device.

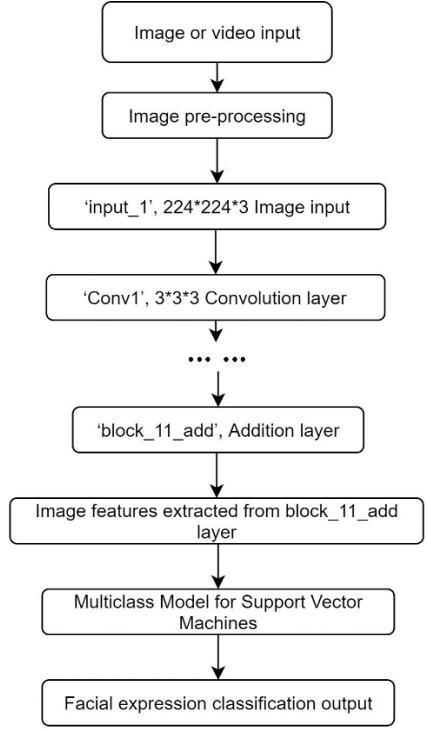

**Fig. 2.** Structure of the emotion recognition algorithm

Depthwise Separable Convolutions are important features in MobileNet. Assume the convolution layer has the $h_i \times w_i \times c_i$ input feature map $I$ where $h_i, w_i, c_i$ represent the height and width for the input feature map and input channel number and the $h_o \times w_o \times c_o$ output feature map $O$ where $h_o, w_o, c_o$ represent the height and width for the output feature map and output channel number [16]. The size of convolution kernel R is denoted as $h_R \times w_R \times c_i \times c_o$ where $h_R, w_R$ represent the height and width of the kernel. Assuming striding one and padding, the output feature map for the standard convolution is represented as:

$$O_{R,l,c_o} = \sum_{d,j,c_i} R_{d,j,c_i,c_o} \cdot I_{R+d-1,l+j-1,c_i} \tag{1}$$

Also, the computational cost for the standard convolution $M_c$ can be represented by the equation where k is the size of the convolution core:

$$M_c = k^2 c_i c_o \times h_o w_o \tag{2}$$

The Depthwise Separable Convolution is the sum of the depthwise convolution and the pointwise convolution. The computational cost for the depthwise convolution $M_d$, the pointwise convolution $M_p$ and the Depthwise Separable Convolution $M_s$ can be represented by the following equation respectively [16]:

$$M_d = k^2 c_i \times h_o w_o \tag{3}$$

$$M_p = c_i c_o \times h_o w_o \tag{4}$$

$$M_s = k^2 c_i \times h_o w_o + c_i c_o \times h_o w_o \tag{5}$$

The computational cost for the Depthwise Separable Convolution and the standard convolution can be compared with the following equations [16]:

$$\frac{M_s}{M_c} = \frac{k^2 c_i \times h_o w_o + c_i c_o \times h_o w_o}{k^2 c_i c_o \times h_o w_o} \tag{6}$$

$$\frac{M_s}{M_c} = \frac{1}{c_o} + \frac{1}{k^2} \qquad (7)$$

Finally, the SVM is utilized to classify the input images in order that the recognition of emotions is achieved. In the research, deep convolution networks are used to extract deep features from the images. The deep features will act as the input for the traditional classifier SVM. The trained SVM is used to recognize facial expressions.

In the convolution networks such as AlexNet and MobileNetV2, softmax classifiers are used to classifier the image features. Also, different classifiers may have different classification performance and characteristics. For example, SVM can be applied to face recognition and object classification. SVM can change many feature representations into higher dimensional spaces by using kernels and it can make a classification in multiple classes [18]. On the other hand, LDA is able to find the optimal transformation to make classification [26]. LDA can be applied to image retrieval and object classification.

*2.3 Evolution of Emotions*

In the experiment, the MoCA score is used to work as the ground truth for group division for cognitively impaired and cognitively healthy participants: Participants who have MoCA scores between 25 and 30 are regarded as cognitively healthy participants and participants who have the MoCA score between 20 and 24 are considered as having the mild cognitive impairment and they belong to the cognitively impaired participant group.

In the following paragraph, how to obtain the data matrix will be introduced. The video of facial expression from a participant can be divided into continuous video frames. It is defined that $n$ is the number of video frames in the video. It is assumed at each time interval $t_n$, there is a stimulus $S_n$ and a participant has an emotion $E_n$ which can be represented by the following equations:

$$t = \{t_1, t_2, t_3, \ldots, t_n\} \qquad (8)$$

$$S = \{S_1, S_2, S_3, \ldots, S_n\} \qquad (9)$$

$$\bar{E} = \{E_1, E_2, E_3, \ldots, E_n\} \qquad (10)$$

It is defined that the participants may express one of the 6 emotions at any time internal: sad, surprise, happy, neutral, angry and other. The emotion of the participants in each frame can be recognized by the proposed facial expression recognition system. In addition, the emotion situation of a participant can be represented by the following matrix using possibilities of six emotions:

$$E_n = \begin{pmatrix} E_{1n} \\ E_{2n} \\ E_{3n} \\ E_{4n} \\ E_{5n} \\ E_{6n} \end{pmatrix} \qquad (11)$$

Furthermore, the evolution of emotions in a period of time for one participant can be defined as the following matrix:

$$E = \begin{bmatrix} E_{11} & E_{12} & \ldots & E_{1n} \\ E_{21} & E_{22} & \ldots & E_{2n} \\ E_{31} & E_{32} & \ldots & E_{3n} \\ E_{41} & E_{42} & \ldots & E_{4n} \\ E_{51} & E_{52} & \ldots & E_{5n} \\ E_{61} & E_{62} & \ldots & E_{6n} \end{bmatrix} \qquad (12)$$

As a result, by applying the proposed facial expression recognition algorithm and data processing, a video of facial expression from a participant can be converted into a matrix $E$ showing the evolution of emotions. As introduced in Fig. 1., the data matrix *E* will work as the input for the SVM classifier and it will be used to visualize the evolution of facial emotions.

## 3. Experiments and Results

*3.1 Introduction to the Experiment*

In this section, experiments are used to test the performance of several facial expression recognition algorithms for recognition accuracy. These algorithms are including the deep learning algorithm AlexNet, one algorithm from the author's previous work [24] and 4 new developed algorithms which will be introduced in the following paragraphs. Also, these algorithms will be compared using 3 facial expression datasets.

First of all, the pre-trained deep convolution network AlexNet is trained with minimum batch size 5, initial rate 0.0003 and maximum epochs 10 [25]. As discussed in the previous paragraph, the methods which combined different deep convolution networks with traditional classifier SVM are compared [27]. In the experiment, different deep convolution networks including AlexNet, MobileNet and ShuffleNet are used to extract features from the images. There are some Hyperparameters involved in this process. Experiments have been conducted to find the optimal hyperparameters using the 'OptimizeHyperparameters' function in Matlab. The function is used to optimize the experiment results by changing the hyperparameters Delta and Gamma automatically. In the experiment, it was found that the operating time was increased by 100 times and there was only little improvement in recognition accuracy for the SVM classifier. The disadvantage of increasing operating time greatly was found which was not appropriate in the current system. As a result, the default hyperparameters for the SVM classifier is used. Furthermore, the performance of facial expression recognition is affected by the layers to extract the features from the deep convolution networks. In summary, for the combination of deep convolution networks and SVM, the following layers to extract the features are used: FC6 layer from AlexNet, block_11_add and block_15_add layers from MobileNet, node_159 and node_163 layers from ShuffleNet.

In order to test the performance of facial expression recognition, several datasets are selected to test and compare different methods. These methods are tested in two kinds of datasets: online datasets like the KDEF dataset [28] and datasets developed from the experiment using the proposed system including the Chinese Adults Dataset and the Chinese Elderly People Dataset.

**Table 1** Summary for 3 Facial Expressions Datasets in Experiments

| Dataset | Number of different emotions | Total number of images selected | Type of emotions | Number of images in each emotion category | Source |
|---|---|---|---|---|---|
| KDEF | 7 | 980 | Lab-posed emotion | Even | Online [28] |
| Chinese Adults Dataset | 5 | 165 | Naturally expressed emotion during experiments | Uneven | Experiment for the proposed system |
| Chinese Elderly People Dataset | 6 | 13692 | Naturally expressed emotion during experiments | Uneven | Experiment for the proposed system |

Table 1 is a summary of all the datasets used in the experiments. The first column shows the name of the three datasets. The second and third columns show the number of types of emotions involved in the dataset and the number of images in the dataset respectively which will be discussed in detail in Part 3.2. The fourth column shows the types how the emotions are expressed: naturally expressed emotions and lab-posed emotions. In the datasets of lab-posed emotions, the actors were trained to express the same emotion in the same way. The images of emotions were also taken in the same sites to ensure the same background, viewpoint and lighting situations. For the datasets of naturally expressed emotion, people may express their emotions in different ways. For instance, the AffectNet dataset contains the images of emotions expressed naturally by normal people [29]. The images of facial expressions are collected from the Internet using 1250 facial expressions-related keywords by using three major search engines [29]. As a result, the images of emotions are in different lighting situations, viewpoints and background environments.

The fifth column is about whether the number of images in different emotion types is even or not. In the datasets with lab-posed emotions, the number of the images can be even. However, in practical situations, when people express their emotions naturally when using the proposed system, they may express more neutral and happy emotions and very few disgust and fear emotions. As a result, in the datasets with the emotions expressed naturally, there are more samples in some emotion categories and fewer samples in some other emotion categories so that the difficulty of recognition of different emotions are different. Finally, the sixth column is about the source of images in the datasets whether they are online datasets or they are developed by the author from the experiment to test the proposed system. In addition, in the designed experiment setting, only front-view faces are involved and the images of side-view faces are removed from these datasets.

In the experiments, it is noticed that some methods may have varied emotion recognition performance in different datasets. There are several reasons. First of all, some factors such as lighting situations, background environment and the viewpoint of the faces may affect the algorithm to recognize the emotions. Some methods may have poor performance in a complex background environment, but the lighting situations may have less effect on these methods. The second reason can be the random factor in the experiment. For instance, in an experiment, 50% of the images are selected as the training datasets. Most of these images in the training datasets are from elderly people. The trained network may be good at recognizing facial expressions from elderly people, but it may be poor at recognizing facial expressions from children. As a result, these randomly made datasets may affect the performance of facial expression recognition. On the other hand, in the lab posed datasets, as there are fewer differences between the images in the same emotion categories, the random factor has less effect on the performance of emotion recognition.

*3.2 Experiment results on three datasets*

*3.2.1 KDEF Dataset Experiments*

**Table 2** Comparison of Recognition Accuracy and Cross-Validation Error Rate Using Different Methods for the KDEF Dataset

| Method | 1 | 2 | 3 | 4 | 5 | Average Recognition Accuracy (%) | Error Rate (%) |
|---|---|---|---|---|---|---|---|
| AlexNet | 84.5 | 83.3 | 82.0 | 82.0 | 83.2 | 83.0 | 20.4 |
| AlexNet + FC6 + SVM | 84.1 | 85.7 | 84.5 | 86.9 | 86.1 | 85.5 | 14.0 |
| MobileNet + block_11_add + SVM | 88.6 | 86.1 | 91.4 | 89.8 | 87.4 | 88.7 | 11.1 |
| MobileNet + block_15_add + SVM | 85.3 | 85.3 | 91.0 | 86.1 | 86.5 | 86.8 | 14.3 |
| ShuffleNet + node_159 + SVM | 87.8 | 89.8 | 86.5 | 88.2 | 85.7 | 87.6 | 13.3 |

| ShuffleNet + node_163 + SVM | 88.6 | 89.4 | 86.9 | 87.4 | 86.9 | 87.8 | 12.0 |

In the KDEF dataset, there are 4900 images from both male and female actors in 5 different angles and in the resolution of 562 × 762. These images of facial expressions are posed by 35 male actors and 35 female actors. In the experiment, 980 front-view facial expression images were selected. Image processing techniques were used to process these images. All the side-view images were removed as the front-view images were focused in the experiment. There are 7 different facial expressions included in the images: neutral, fear, angry, surprise, happy, disgust and sad. To examine each of these methods, 75% of images formed the training dataset and all the other images formed the testing dataset. These training images were randomly selected and they were changed each time. Table 2 shows the recognition accuracy using images from the KDEF dataset with different methods. The selected methods are listed in the first column. The experiment results of five times are shown from column two to column six. Also, the average facial expression recognition accuracy is shown in column seven. For validation, the 5-fold cross-validation error rate is shown in column eight.

As shown in Table 2, the proposed method (MobileNet + block_11_add + SVM) has the best facial expression recognition performance which is 88.7% and is better than other combinations on average. Also, the proposed method shows the lowest error rate. The deep convolution network AlexNet only has a recognition accuracy of 83.0% which is much lower than the proposed method.

In addition, it is noticed that the proposed method (MobileNet + block_11_add + SVM) has the best performance four times and has the third best performance for one time. The major reason is from the random factor which has been explained in the previous section. The variations of recognition accuracy are relatively high each time as there are only 980 images selected in the dataset.

*3.2.2 Chinese Adults Dataset Experiments*

**Table 3** Comparison of Recognition Accuracy and Cross-Validation Error Rate Using Different Methods for the Chinese Adults Dataset

| Method | 1 | 2 | 3 | 4 | 5 | Average Recognition Accuracy (%) | Error Rate (%) |
|---|---|---|---|---|---|---|---|
| AlexNet | 61.9 | 50.0 | 66.7 | 69.1 | 71.4 | 63.8 | 43.7 |
| AlexNet + FC6 + SVM | 66.7 | 69.1 | 73.2 | 71.4 | 66.7 | 69.4 | 33.3 |
| MobileNet + block_11_add + SVM | 80.9 | 71.4 | 66.7 | 71.4 | 73.8 | 72.8 | 26.0 |
| MobileNet + block_15_add + SVM | 71.4 | 69.1 | 69.1 | 69.1 | 73.8 | 70.5 | 27.3 |
| ShuffleNet + node_159 + SVM | 76.2 | 66.7 | 64.3 | 78.6 | 71.4 | 71.4 | 31.5 |
| ShuffleNet + node_163 + SVM | 76.2 | 71.4 | 64.3 | 76.2 | 73.8 | 72.4 | 32.1 |

This dataset contains 165 images of facial expressions from Chinese adults when they are using the proposed system. The experiment was done in Shanghai Mental Health Center. The images are processed using the image pre-processing techniques. The ground truth of the emotions is labelled by the first author. This dataset contains images of emotions expressed naturally from both male and female participants when they are watching the video stimuli. The images are resized to 227 × 227. The dataset involves five different emotions: happy, surprise, angry, neutral and sad. Emotions of fear and disgust are not included in the dataset as the participants show very few these emotions during the experiment. 75% of the images are selected randomly as the training images and the remaining images are working as the testing images.

Table 3 compares the recognition performance using the Chinese Adults Dataset with different methods. The first column shows different methods of emotion recognition. From column two to column six, recognition accuracies in five times using different methods are shown. Column seven states the average facial expression recognition accuracy using different methods. Also, column eight shows the 5-fold cross-validation error rate.

The proposed method (MobileNet + block_11_add + SVM) also has the best facial expression recognition accuracy on average which is 72.8% and the lowest error rate which is 26.0%. As the subsets are randomly selected in 5-fold cross-validation, there is a small difference between the average recognition accuracy and cross-validation error rate. AlexNet has worse recognition accuracy which is 63.8% than the other methods which combine deep convolution networks and SVM classifiers.

Compared to other datasets, the average recognition accuracy for all methods is about 70% which is quite low. There are several factors which cause the low recognition performance. To start with, this dataset only contains 165 images of facial expressions from Chinese adults and the classifier is hard to learn the features of each emotion type well. Also, some emotions are not obvious and the difference between each emotion type is small. In addition, the small number of images in the dataset also causes large variations of recognition accuracy each time.

*3.2.3 Chinese Elderly People Dataset Experiments*

**Table 4** Comparison of Recognition Accuracy and Cross-Validation Error Rate Using Different Methods for the Chinese Elderly People Dataset

| Method | 1 | 2 | 3 | 4 | 5 | Average Recognition Accuracy (%) | Error Rate (%) |
|---|---|---|---|---|---|---|---|
| AlexNet | 93.6 | 92.9 | 92.9 | 92.7 | 93.0 | 93.0 | 6.9 |
| AlexNet + FC6 + SVM | 95.5 | 95.0 | 95.6 | 95.5 | 95.5 | 95.4 | 4.4 |
| MobileNet + block_11_add + SVM | 97.6 | 97.8 | 97.2 | 97.6 | 97.6 | 97.6 | 2.7 |
| MobileNet + block_15_add + SVM | 96.8 | 96.7 | 96.7 | 96.8 | 96.8 | 96.8 | 3.3 |
| ShuffleNet + node_159 + SVM | 97.0 | 96.5 | 96.8 | 97.0 | 97.0 | 96.9 | 3.4 |
| ShuffleNet + node_163 + SVM | 97.2 | 96.8 | 96.9 | 97.2 | 97.2 | 97.1 | 2.9 |

As the system is mainly aimed to detect the cognitive impairment for elderly people, most participants are Chinese elderly people including both male and female participants in the experiment. A larger dataset containing 13692 images is set up using facial expressions from Chinese elderly people when they are watching the video stimuli through the proposed system. The experiment was also done in Shanghai Mental Health Center. The image pre-processing technique is used to process the images and the first author labelled the ground truth of emotion type for each image manually.

There are six different emotion types in the dataset: happy, surprise, angry, neutral, sad and other. This dataset doesn't contain the emotions of disgust and fear, but the other emotion is included in the dataset when the participant is distracted by other things or his or her face is coved by obstacles. 75% of the images are used as the training images randomly and the rest images are used as the testing images.

Table 4 compares the recognition accuracy using the Chinese Elderly People Dataset with different methods. Different methods for emotion recognition are listed in the first column. Furthermore, recognition accuracies in five times using different methods are listed from column two to column six. Column seven shows the average facial expression recognition accuracy using different methods. In final, the 5-fold cross-validation error rate is shown in column eight.

Table 4 shows that the proposed method (MobileNet + block_11_add + SVM) also has the best facial expression recognition accuracy on average which is 97.6% and the lowest error rate which is 2.7%. All the algorithms have good performance in this dataset, as there are 13692 images in the dataset and let the classifier learn the features of different emotion types well. Also, the large number of images in the dataset results in low variations in recognition accuracy in five times and similar results for average recognition accuracy and cross-validation error rate.

*3.3 Emotion Patterns between Normal People and Patient Comparison*

In the experiment, the videos of facial expressions are analyzed to find the difference in emotion patterns between cognitively impaired people and cognitively healthy people. In the experiment, the occurrence of facial expressions is used to compare their different emotion patterns. For instance, at a select time, we can compare the occurrence of happiness between the group of cognitively impaired participants and cognitively healthy participants and we may find if the cognitively impaired participants show more happy emotion on average or not. It is defined that the occurrence of one type of emotion U is obtained by:

$$U = \frac{N}{T} \qquad (13)$$

Where $N$ stands for the number of people who show this type of emotion, $T$ is the total number of people in the group. In addition, the evolution of happy emotion in a participant group in a period of time can be found by finding the occurrence of this emotion at every frame in the video. In the experiment, it is found that the major difference between cognitively impaired people and cognitively healthy people are in the emotion of happy, sad and angry and there is a small difference in the emotion of neutral and other. The major difference will be shown in the following figures, where the *x*-axis stands for the frame number and the y-axis stands for the occurrence of one emotion.

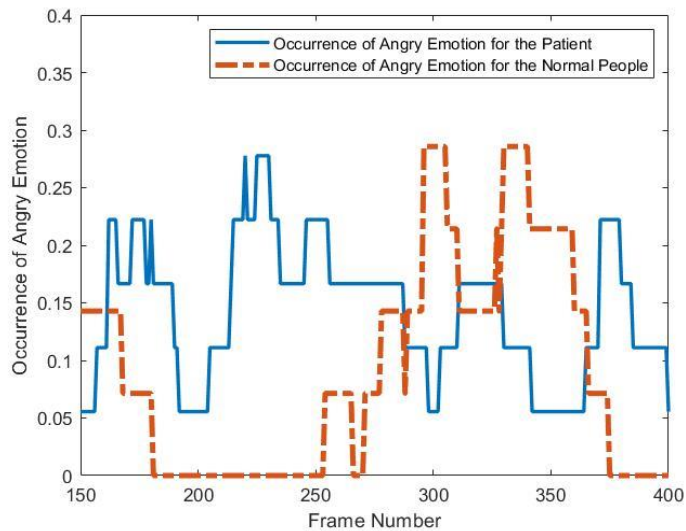

**Fig. 3.** Evolution of angry emotion

Fig. 3. mainly shows the different emotion patterns in angry emotion between cognitively impaired people and cognitively healthy people. The figure shows a large difference in angry emotion between these two groups of people. Especially from frame 200 to frame 250, the cognitively impaired

participants group show more angry emotion in this time period which suggests that the patients are more likely to be affected by the video stimuli at that time.

In addition, the difference in evolution pattern of happy emotion in the experiment is shown in Fig. 4. between cognitively impaired people and healthy people. This part of the experiment shows that normal people show much more happy emotion during watching the video stimuli than cognitively impaired people.

Finally, Fig. 5. mainly shows the difference in the sad emotion. It is shown that the cognitively impaired people show less sad emotion from frame 220 to frame 280 compared to the cognitively healthy people.

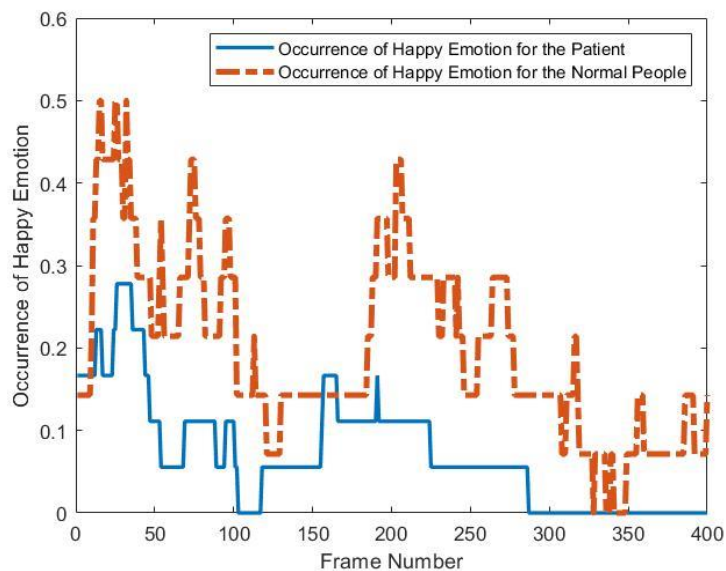

**Fig. 4.** Evolution of happy emotion

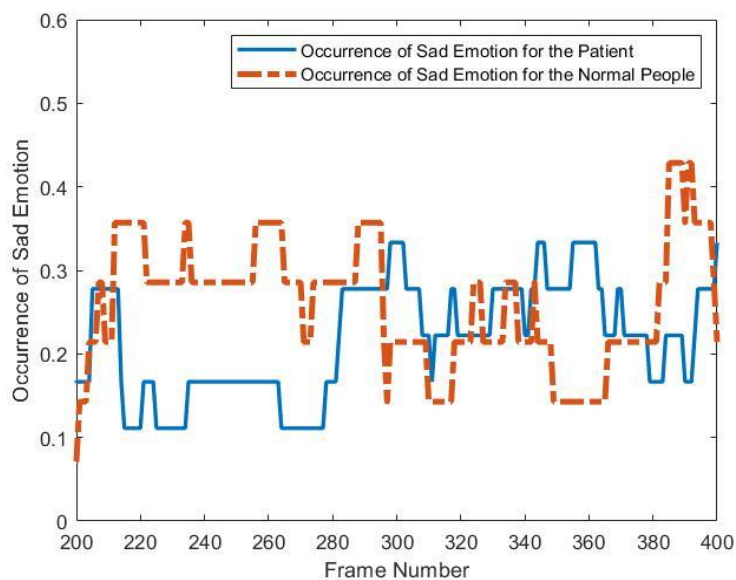

**Fig. 5.** Evolution of sad emotion

*3.4 Proposed System to Detect the Cognitive Impairment*

As introduced in Section 2.1, after the videos of participants are recognized by the proposed algorithm, the SVM classifier can be used to detect the cognitive impairment which will be explained in detail.

As it is introduced in equation (12) in Section 2.4, the evolution of facial expressions of a participant when watching the video can be represented by a matrix $E$. Next, the data selection process will only choose some video frames and only 5 emotions. Among the 6 emotions including happy, angry, sad, neutral, other and surprise, the surprise emotion is excluded, as few participants show the surprise emotion when watching the video stimuli. Also, the time period for emotion is selected when there is the largest difference between the cognitively impaired participants and cognitively healthy participants.

**Table 5** Comparison of Cognitive Impairment Detection Accuracy Using Different Classification Methods

| Method | Detection Accuracy (%) |
|---|---|
| LDA | 60.0 |
| SVM | 73.3 |
| KNN | 60.0 |
| Decision Tree | 40.0 |

In the experiment, the dataset includes emotion data from 36 cognitively impaired people and 25 cognitively healthy people. In addition, the training dataset contains the emotion data from 46 participants including 18 healthy people and 28 cognitively impaired people and the testing dataset contains the emotion data from the remaining 15 participants including 7 healthy people and 8 cognitively impaired people. Classifiers such as LDA, SVM, KNN and decision tree classifiers are compared in this experiment. These classifiers are trained in the training dataset to learn the emotion pattern from cognitively impaired people. Then, the trained classifiers are used to detect the cognitive impairment in the testing dataset. As shown in Table 5, the SVM classifier achieves the best result of 73.3% in detecting cognitive impairment.

**4 Discussion**

In the experiment, 6 facial expression recognition algorithms are compared using 3 datasets, including deep convolution network AlexNet and algorithms combining deep convolution networks with traditional classifiers. The comparison results are mainly shown in Table 6. In this table, 6 facial expression recognition algorithms are listed in column one and from column two to column four, the recognition performance is compared among 6 methods using 3 datasets.

It is noticed that the proposed method (MobileNet + block_11_add + SVM) has the best emotion recognition performance in all these three datasets. In addition, by combining deep convolution networks and traditional classifiers, the performance is better than AlexNet in all these methods.

**Table 6** Comparison of Emotion Recognition Accuracy Using Different Methods for Three Datasets

| Method | KDEF | Chinese Adults | Chinese Elderly People |
|---|---|---|---|
| AlexNet | 83.0 | 63.8 | 93.0 |
| AlexNet + FC6 + SVM | 85.5 | 69.4 | 95.4 |
| MobileNet + block_11_add + SVM | 88.7 | 72.8 | 97.6 |
| MobileNet + block_15_add + SVM | 86.8 | 70.5 | 96.8 |
| ShuffleNet + node_159 + SVM | 87.6 | 71.4 | 96.9 |
| ShuffleNet + node_163 + SVM | 87.8 | 72.4 | 97.1 |

In addition, the proposed method (MobileNet + block_11_add + SVM) has the best performance in the KDEF dataset which is 88.7% compared to the performance of other researchers' algorithms which is shown in Table 7. On the other hand, it is noticed that each method from different published papers has their experiment settings which may cause a difference in recognition performance. As stated in Section 3.2, 75% of the images are used as training datasets randomly.

**Table 7** Comparison of Recognition Performance from Other Published Papers

| Method | KDEF (%) |
|---|---|
| DeepPCA[30] | 83.0 |
| AAM+SVM[31] | 74.6 |
| Feature+SVM[32] | 82.4 |
| Proposed method | 88.7 |

In practical situations, operating time is important as well as recognition accuracy. Fig. 6. mainly compares the operating time and recognition accuracy for the proposed method (MobileNet + block_11_add + SVM) and some state-of-the-art deep learning methods including AlexNet, MobileNet, GoogleNet and ShuffleNet in KDEF dataset. The overall operating time and training time are represented by the line chart and the recognition accuracy is represented by the clustered columns. It is noticed that the proposed method has the shortest training time and overall operating time and also has the best recognition accuracy. It is noticed that the proposed method has about 13 times shorter training time and overall operating time than ShuffleNet which is important in the proposed cognitive impairment detection system.

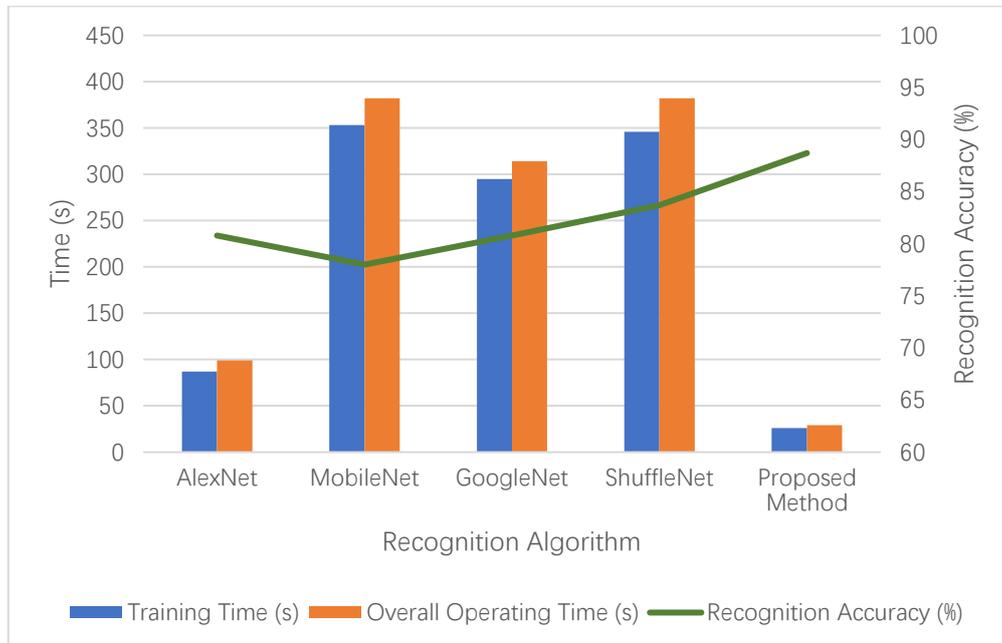

**Fig. 6.** Comparison of training time, overall operating time and recognition accuracy for different methods in KDEF dataset

**5 Conclusion**

In conclusion, a novel cognitive impairment detection system has been proposed based on facial expression recognition with acceptable accuracy and low cost, which can detect the cognitive impairment in the early stage. The cognitive impairment detection system was tested in Shanghai Mental Health Center in 2019. 61 participants took part in the experiments to verify the proposed system. The proposed system successfully collected their emotion data when they were watching designed video stimuli. In addition, a dataset was built including both cognitively impaired participants and cognitively healthy participants. The proposed facial expression recognition algorithm (MobileNet + block_11_add + SVM) was used to recognize the facial expressions from the videos. After that, 61 pieces of data about the evolution of emotions were obtained after data processing. After being trained with 46 pieces of emotion data, a SVM classifier was able to detect the cognitive impairment based on the emotion data from the testing dataset with a detection accuracy of 73.3%.

Although the proposed algorithm has good recognition accuracy and short operating time, it may have problems in recognizing minor emotions. In the experiment, the participants may show some minor emotions naturally when watching the designed video stimuli. There is only a small difference between the minor emotions and the neutral emotion. These minor emotions may be recognized as neutral emotion wrongly. To solve this problem, the algorithm needs to focus on some specific facial parts in the face. Also, in order to use the system to detect the cognitive impairment for elderly people with better performance, the proposed algorithm needs to be trained with large amounts of images of natural emotions from elderly people. Moreover, the emotion recognition accuracy can be further improved and the operating time can be reduced.

For future work, the developed system can operate on mobile devices as an app. The system helps to detect the cognitive impairment for the users while they are watching the designed stimuli through the mobile phones. As a result, this app can work as a low-cost cognitive impairment detection solution for elderly people all over the world.

**Conflicts of Interest:** The authors declare no conflict of interest.

**Acknowledgements:** Huiyu Zhou was partly funded by Royal Society-Newton Advanced Fellowship under Grant NA160342. The authors thank Shanghai Mental Health Center for their help and support.

**Ethics:** The authors declare that all procedures were performed in compliance with relevant laws and institutional guidelines. The authors declare that the appropriate institutional committee(s) has approved them. The authors declare that informed consent was obtained for experimentation with human subjects.